\documentclass[runningheads]{llncs}
\usepackage[T1]{fontenc}
\usepackage{graphicx,verbatim}
\usepackage{amsmath}
\usepackage{amssymb}
\usepackage{bbm}

\begin{document}

\title{PelviNeXt: A Modality-Agnostic Hybrid Network for Pelvic Imaging in Women's Health}
\titlerunning{PelviNeXt}

\author{Siam Tahsin Bhuiyan\inst{1, 2}\orcidID{0009-0000-1298-1991}, Rashedur Rahman\inst{1, 2}\thanks{Corresponding author}\orcidID{0000-0003-0267-2612}, Sefatul Wasi\inst{2}\orcidID{0009-0004-6949-0607}, Halima Khatun\inst{1, 2}\orcidID{0009-0008-9022-4441}, Ashraful Islam\inst{1, 2}\orcidID{0000-0003-2367-2013}, AKM Mahbubur Rahman\inst{1, 2}\orcidID{0000-0001-9941-4817}, Saadia Binte Alam\inst{1, 2}\orcidID{0009-0007-0358-7635}, M Ashraful Amin\inst{1, 2}\orcidID{0000-0003-2330-9775}} 
\authorrunning{S. T. Bhuiyan et al.}
\institute{Center for Computational \& Data Sciences, Independent University, Bangladesh, Dhaka, Bangladesh \and
Department of Computer Science and Engineering, Independent University, Bangladesh, Dhaka, Bangladesh \\
\email{rashed@iub.edu.bd}}

\maketitle

\begin{center}
\small\itshape
Preprint. Accepted at MICCAI CAPI-WOMEN 2026.
\end{center}

\vspace{0.5em}

\begin{abstract}
Women's health remains substantially under-resourced in medical imaging research,
with pelvic pathologies such as polycystic ovary syndrome (PCOS) and pelvic
fracture both suffering from a scarcity of public, well-annotated benchmark data
despite their clinical importance. We introduce PelviNeXt, a modality-agnostic
hybrid architecture combining a dense convolutional feature extractor,
hierarchical channel-spatial attention (H-CBAM), a multi-scale fusion module
(MSFM), and talking-heads multi-head self-attention (TH-MHSA), applied without
modification to both pelvic ultrasound and X-ray inputs. While benchmarking
PelviNeXt on PCOSGen, the only gynaecologist-annotated public PCOS ultrasound
dataset, we identified extensive exact and near-duplicate contamination within
and across the dataset. We audit this contamination via perceptual hashing,
publicly release a deduplicated version of the dataset, and establish the first
integrity-audited evaluation protocol and baseline for PCOSGen under 5-fold
cross-validation. On the only publicly available pelvic fracture X-ray dataset
(PXR150), PelviNeXt exceeds previously reported state-of-the-art results across
accuracy, recall, specificity, and AUROC. Ablation studies confirm that each
architectural component contributes to performance on both tasks. Our results
demonstrate that a single architecture, applied without task-specific
modification, can serve as a reliable foundation for pelvic imaging across
modalities in data-scarce, under-researched areas of women's health.
\end{abstract}

\section{Introduction}

Despite growing awareness of health disparities in medicine, women's health
remains chronically underfunded and under-researched. Only 1\% of research and
development funding focuses on non-cancer-related women's health, despite women
comprising roughly 50\% of the global population~\cite{moley2024closing}. The
World Economic Forum's 2025 white paper \textit{Prescription for Change},
further reveals that only 7\% of pharmaceutical research and development is
targeted at conditions that uniquely affect women~\cite{bode2026prescribing}.
Left unaddressed, such gaps in research and data risk carrying over into the AI
systems built on top of them.

This gap extends to conditions that disproportionately affect women or share
anatomy with reproductive organs. PCOS exemplifies the former: it affects 6 to
20\% of premenopausal women worldwide~\cite{lv2022deep} and is conventionally
screened via ultrasound. Pelvic fracture exemplifies the latter: though not
sex-specific, it occurs in the anatomical region housing the reproductive
organs, and carries a disability rate exceeding 50\% and a mortality rate above
13\%~\cite{sang2026benchmark}. Plain X-ray remains the primary screening
modality in emergency settings.

Both tasks suffer from scarce public benchmark data. PCOSGen is, to our
knowledge, the only gynaecologist-annotated public PCOS ultrasound dataset,
comprising 4,668 images~\cite{handa_2025_14592001,handa_2025_14591782,handa2024auto}, and has been benchmarked by several pipelines reporting over
96.12\% accuracy~\cite{moral2024cystnet}. Pelvic fracture detection has
likewise reported strong results on private data, e.g. 98.5\% accuracy on 876
radiographs~\cite{kassem2023explainable}, but such data is typically
institution-specific and unreleased. PXR150 is, to our knowledge, the only
public dataset for this task, comprising 150 radiographs (100 fracture, 50
normal)~\cite{cheng2021scalable}.

Existing architectures for these tasks are largely single-modality. Yet both
share structure, namely localized pathological cues within a broader anatomical
context, well suited to dense convolutional backbones~\cite{huang2017densely},
convolutional block attention module (CBAM)-style channel-spatial
attention~\cite{woo2018cbam}, and global self-attention refinements such as
talking-heads attention~\cite{shazeer2020talking}.

We introduce PelviNeXt, a modality-agnostic hybrid architecture combining a
dense convolutional feature extractor, hierarchical channel-spatial attention,
multi-scale fusion, and talking-heads self-attention, applied without
modification across ultrasound and X-ray. While benchmarking PelviNeXt, we
identified extensive exact and near-duplicate contamination in PCOSGen. We
audit this contamination via perceptual hashing and release a deduplicated
dataset, establishing the first integrity-audited evaluation protocol (see Data
Availability).

Our contributions are:
\begin{enumerate}
    \item PelviNeXt, a modality-agnostic architecture evaluated on pelvic
    fracture X-ray and PCOS ultrasound classification with no task-specific
    preprocessing.
    \item A systematic integrity audit of PCOSGen via perceptual hashing,
    revealing extensive near-duplicate contamination, with a deduplicated
    dataset release.
    \item The first reliable, deduplication-aware PCOSGen baseline under
    5-fold cross-validation.
    \item State-of-the-art results on PXR150, exceeding prior work on
    accuracy, recall, specificity, and AUROC.
\end{enumerate}

\section{Methodology}

\begin{figure}[t]
\centering
\includegraphics[width=\textwidth]{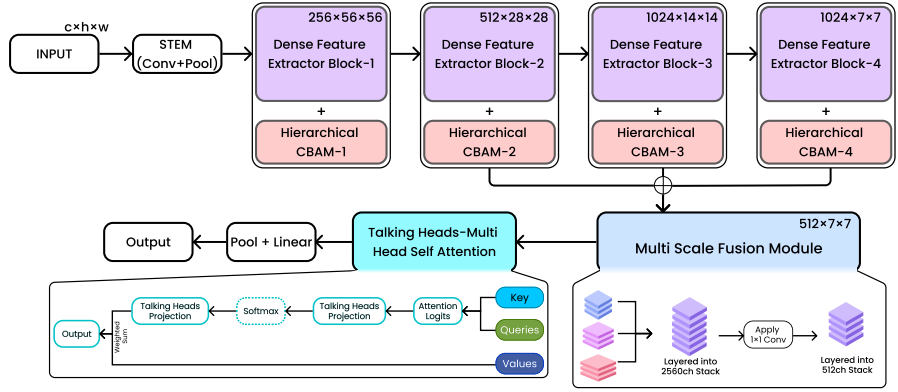}
\caption{Overview of the PelviNeXt architecture.}
\label{fig:architecture}
\end{figure}

PelviNeXt comprises four stages: a dense feature extractor (DFE), hierarchical
CBAM (H-CBAM) attention after each DFE block, a multi-scale fusion module
(MSFM), and a talking-heads multi-head self-attention (TH-MHSA) module,
followed by a classification head (Fig.~\ref{fig:architecture}). The same
pipeline with identical hyperparameters is applied to both ultrasound and X-ray
inputs.

\subsection{Dense Feature Extractor (DFE)}
The feature extraction stage is inspired by the depth and dense connectivity
of DenseNet~\cite{huang2017densely}, using four dense blocks of 6, 12, 24, and
16 layers respectively (DFE-1 through DFE-4), where each layer receives the
concatenated feature maps of all preceding layers within the block. Each block
is followed by a transition layer halving spatial resolution, producing feature
maps of $256{\times}56{\times}56$, $512{\times}28{\times}28$,
$1024{\times}14{\times}14$, and $1024{\times}7{\times}7$ after DFE-1 through
DFE-4 respectively, given a $3{\times}224{\times}224$ input. 

\subsection{Hierarchical CBAM (H-CBAM)}
Following each DFE block, an H-CBAM module refines the feature map along
channel and spatial dimensions~\cite{woo2018cbam}. Given feature map
$\mathbf{F} \in \mathbb{R}^{C \times H \times W}$, channel attention is
\begin{equation}
M_c(\mathbf{F}) = \sigma\big(\mathrm{MLP}(\mathrm{AvgPool}(\mathbf{F})) +
\mathrm{MLP}(\mathrm{MaxPool}(\mathbf{F}))\big)
\end{equation}
and spatial attention is
\begin{equation}
M_s(\mathbf{F}') = \sigma\big(f^{7\times7}([\mathrm{AvgPool}(\mathbf{F}');\,
\mathrm{MaxPool}(\mathbf{F}')])\big)
\end{equation}
where $\mathbf{F}' = M_c(\mathbf{F}) \otimes \mathbf{F}$, $\sigma$ is sigmoid,
and $f^{7\times7}$ is a $7{\times}7$ convolution. The refined output is
$\mathbf{F}'' = M_s(\mathbf{F}') \otimes \mathbf{F}'$. We instantiate this
module separately after each DFE block (H-CBAM-1 through H-CBAM-4), applying
attention at four progressively coarser scales from $56{\times}56$ down to
$7{\times}7$, enabling refinement of both fine-grained local cues and coarse
semantic structure within a single forward pass.

\subsection{Multi-Scale Fusion Module (MSFM)}
The outputs of DFE-2, DFE-3, and DFE-4 after their respective H-CBAM modules,
$B_2 \in \mathbb{R}^{512\times28\times28}$,
$B_3 \in \mathbb{R}^{1024\times14\times14}$, and
$B_4 \in \mathbb{R}^{1024\times7\times7}$, are spatially aligned to
$7{\times}7$ via bilinear interpolation and concatenated into a 2560-channel
tensor. A $1{\times}1$ convolution reduces this to 512 channels, integrating
mid-level and high-level features before global reasoning.

\subsection{Talking-Heads Multi-Head Self-Attention (TH-MHSA)}
The fused $512{\times}7{\times}7$ map is flattened into 49 spatial tokens of
dimension 512 and processed by a talking-heads self-attention
block~\cite{shazeer2020talking}, which inserts learned linear projections across
the head dimension before and after the softmax:
\begin{equation}
\mathrm{Attention}(Q,K,V) = \mathrm{softmax}\!\left(\frac{QK^\top}{\sqrt{d_k}}
W_\ell\right) W_w\, V
\end{equation}
where $W_\ell$ and $W_w$ mix information across heads pre- and post-softmax,
improving global context aggregation relative to standard multi-head
self-attention. Output tokens are mean-pooled and passed through a linear layer
to produce class logits.

\section{Datasets and Experimentation}

\subsection{PCOSGen Dataset}
PCOSGen is the dataset released for the Auto-PCOS Classification
Challenge~\cite{handa2024auto}, collected from YouTube, ultrasoundcases.info,
and Kaggle, and annotated by an experienced gynaecologist based in New Delhi,
India~\cite{handa_2025_14592001,handa_2025_14591782}. The separate training and
test releases (PCOSGen-train: 3,200 images; PCOSGen-test: 1,468 images) are
merged into a single pool of 4,668 images (1,319 Normal, 3,349 Abnormal) as no
official train/test boundary is assumed in our 5-fold cross-validation protocol.

\subsection{PCOSGen Integrity Audit}
\begin{figure}[t]
\centering
\includegraphics[width=\textwidth]{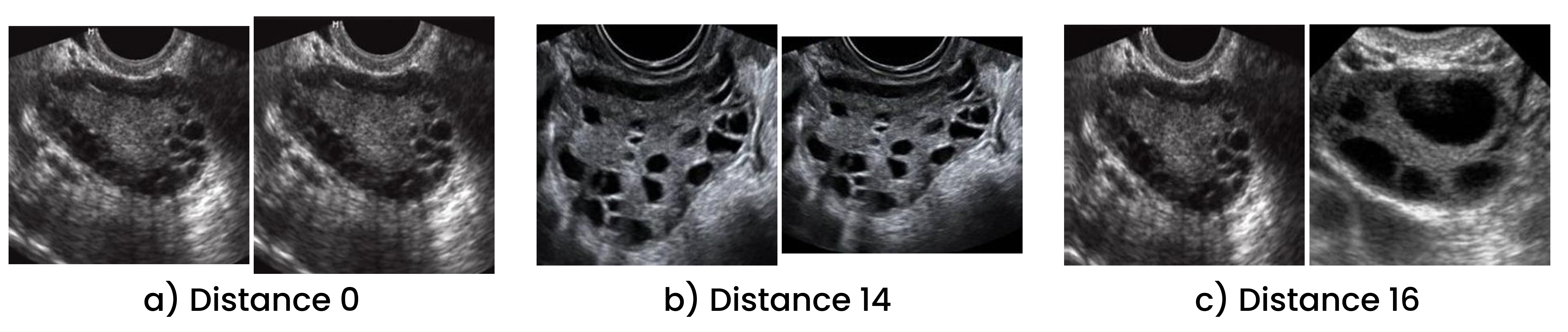}
\caption{Example image pairs at increasing perceptual hash distance. (a)
Distance 0: exact duplicate. (b) Distance 14: visually indistinguishable,
used as the deduplication threshold. (c) Distance 16: visually distinct,
confirming the threshold boundary.}
\label{fig:dupes}
\end{figure}

We compute a perceptual hash (pHash) for every image and calculate pairwise
Hamming distances. Images connected at or below a given distance threshold are
grouped into clusters via union-find, treating duplication as transitive across
chains of near-identical images. Deduplication proceeds in two stages: exact
duplicates (distance $=0$) are removed first, retaining one representative per
cluster; near-duplicates (distance $\leq 14$) are then removed from the
remaining images. The threshold of 14 was selected by visual inspection: pairs
at distance 14 are visually indistinguishable, while pairs at distance 16 are
visually distinct (Fig.~\ref{fig:dupes}). Table~\ref{tab:audit} summarizes each
stage. The original pool of 4,668 images reduces to 225 (63 Normal, 162
Abnormal), a 95.2\% reduction. The deduplicated dataset is publicly released
(see Data Availability).

\begin{table}[t]
\footnotesize
\setlength{\tabcolsep}{3.5pt}
\renewcommand{\arraystretch}{0.92}
\centering
\caption{PCOSGen dataset composition before and after deduplication.}
\label{tab:audit}
\begin{tabular*}{\columnwidth}{@{\extracolsep{\fill}}lccc@{}}
\hline
\textbf{Dataset} & \textbf{Normal} & \textbf{Abnormal} & \textbf{Total} \\
\hline
Original PCOSGen & 1{,}319 & 3{,}349 & 4{,}668 \\
After exact duplicate removal & 948 & 2{,}523 & 3{,}471 \\
After near-duplicate removal ($\leq 14$) & 63 & 162 & 225 \\
\hline
\end{tabular*}
\end{table}

\subsection{Pelvic Fracture Dataset (PXR150)}
PXR150 is a publicly available 150-image subset of the test set from
\cite{cheng2021scalable}, originally divided into 50 hip fracture, 50 pelvic
fracture, and 50 normal cases from a 2017 emergency department cohort of 1,888
radiographs. For binary classification, the two fracture classes are merged
into a single Fracture class, giving 100 fracture and 50 normal cases.

\subsection{Implementation Details}
All models are trained from scratch using AdamW (lr $= 1{\times}10^{-4}$,
cosine annealing), batch size 16, for 30 epochs, under 5-fold stratified
cross-validation with augmentation applied only to training folds. Augmentation
uses random combinations of rotation (within $25^\circ$), shearing (within
10\%), horizontal flipping, and translation (within 10\%), with class-aware
multipliers: fracture images $2{\times}$ and normal images $4{\times}$ for
PXR150; abnormal images $2{\times}$ and normal images $6{\times}$ for the
deduplicated PCOS dataset. No task-specific preprocessing is applied to either
modality.

\subsection{Evaluation Metrics}
We report Accuracy, Recall, Specificity, F1-Score, and AUROC for both tasks,
with mean and 95\% confidence interval (CI) computed across the 5
cross-validation folds.

\section{Results \& Discussion}

\subsection{PCOS Classification}
Table~\ref{tab:pcos} reports 5-fold CV results on the deduplicated PCOSGen
dataset. PelviNeXt achieves the highest performance across all five metrics,
with a mean accuracy of 92.00\% ($\pm$1.60) and AUROC of 0.9051
($\pm$0.0156), outperforming ViT-B/16~\cite{dosovitskiy2020image},
ResNet-101~\cite{he2016deep}, and DenseNet-169~\cite{huang2017densely} by at
least 2.67 percentage points in accuracy and 0.0422 in AUROC. PelviNeXt also
shows the narrowest CIs across most metrics: specificity varies by only
$\pm$2.99 percentage points for PelviNeXt versus $\pm$12.59 for ViT-B/16,
suggesting more stable performance in this low-data regime. As no prior work
has been evaluated on the deduplicated dataset, these results constitute the
first reliable, integrity-audited baseline for PCOS classification on PCOSGen.

\begin{table}[t]
\footnotesize
\setlength{\tabcolsep}{3.5pt}
\renewcommand{\arraystretch}{0.92}
\centering
\caption{PCOS classification results on the deduplicated PCOSGen dataset. Mean
over 5-fold CV; 95\% CI shown in the row below each model's results.}
\label{tab:pcos}
\begin{tabular*}{\columnwidth}{@{\extracolsep{\fill}}lccccc@{}}
\hline
\textbf{Model} & \textbf{Acc.} & \textbf{Rec.} & \textbf{Spec.} &
\textbf{F1} & \textbf{AUROC} \\
\hline
ViT-B/16 & 89.33\% & 88.92\% & 79.36\% & 0.8337 & 0.8629 \\
 & \footnotesize$\pm$4.22 & \footnotesize$\pm$3.98 & \footnotesize$\pm$12.59
 & \footnotesize$\pm$0.0802 & \footnotesize$\pm$0.0763 \\
ResNet-101 & 88.44\% & 88.60\% & 80.90\% & 0.8427 & 0.8479 \\
 & \footnotesize$\pm$4.85 & \footnotesize$\pm$3.44 & \footnotesize$\pm$7.85
 & \footnotesize$\pm$0.0402 & \footnotesize$\pm$0.0637 \\
DenseNet-169 & 88.89\% & 87.14\% & 81.03\% & 0.8342 & 0.8418 \\
 & \footnotesize$\pm$3.64 & \footnotesize$\pm$11.90 & \footnotesize$\pm$5.81
 & \footnotesize$\pm$0.0715 & \footnotesize$\pm$0.0664 \\
\hline
\textbf{PelviNeXt} & \textbf{92.00\%} & \textbf{91.74\%} & \textbf{86.48\%} &
\textbf{0.8890} & \textbf{0.9051} \\
 & \footnotesize$\pm$1.60 & \footnotesize$\pm$3.91 & \footnotesize$\pm$2.99
 & \footnotesize$\pm$0.0085 & \footnotesize$\pm$0.0156 \\
\hline
\end{tabular*}
\end{table}

\begin{table}[t]
\footnotesize
\setlength{\tabcolsep}{3.5pt}
\renewcommand{\arraystretch}{0.92}
\centering
\caption{Fracture classification results on PXR150. Mean over 5-fold CV; 95\%
CI shown in the row below each of our models' results.}
\label{tab:fracture}
\begin{tabular*}{\columnwidth}{@{\extracolsep{\fill}}lccccc@{}}
\hline
\textbf{Model} & \textbf{Acc.} & \textbf{Rec.} & \textbf{Spec.} &
\textbf{F1} & \textbf{AUROC} \\
\hline
CLAHE~\cite{bhuiyan2025preprocessing} & 80.67\% & 82.00\% & 80.00\%
& -- & 0.8140 \\
Gamma~\cite{bhuiyan2025preprocessing} & 80.67\% & 81.00\% & 81.00\%
& -- & 0.8160 \\
Ensemble~\cite{bhuiyan2025patch} & 80.00\% & 82.00\% & 76.00\%
& -- & 0.7900 \\
Patch Ensemble~\cite{bhuiyan2025patch} & 84.00\% & 87.00\% & 82.00\%
& -- & 0.8700 \\
\hline
ViT-B/16 & 80.67\% & 84.00\% & 79.00\% & 0.8133 & 0.8210 \\
 & \footnotesize$\pm$5.62 & \footnotesize$\pm$3.67 & \footnotesize$\pm$6.50
 & \footnotesize$\pm$0.0459 & \footnotesize$\pm$0.0535 \\
ResNet-101 & 78.00\% & 79.00\% & 74.00\% & 0.7626 & 0.7870 \\
 & \footnotesize$\pm$3.92 & \footnotesize$\pm$5.71 & \footnotesize$\pm$4.80
 & \footnotesize$\pm$0.0395 & \footnotesize$\pm$0.0574 \\
DenseNet-169 & 78.00\% & 78.00\% & 76.00\% & 0.7697 & 0.7960 \\
 & \footnotesize$\pm$3.33 & \footnotesize$\pm$3.92 & \footnotesize$\pm$5.71
 & \footnotesize$\pm$0.0478 & \footnotesize$\pm$0.0424 \\
\hline
\textbf{PelviNeXt} & \textbf{87.33\%} & \textbf{89.00\%} & \textbf{87.00\%} &
\textbf{0.8774} & \textbf{0.8920} \\
 & \footnotesize$\pm$2.45 & \footnotesize$\pm$5.71 & \footnotesize$\pm$3.92
 & \footnotesize$\pm$0.0191 & \footnotesize$\pm$0.0288 \\
\hline
\end{tabular*}
\end{table}

\subsection{Fracture Classification}
Table~\ref{tab:fracture} reports 5-fold CV results on PXR150. PelviNeXt
achieves the highest accuracy (87.33\%, $\pm$2.45), recall (89.00\%,
$\pm$5.71), specificity (87.00\%, $\pm$3.92), and AUROC (0.8920,
$\pm$0.0288) among all methods, exceeding the strongest prior result, Patch
Ensemble~\cite{bhuiyan2025patch} (84.00\% accuracy, 87.00\% recall, 82.00\%
specificity, 0.8700 AUROC), across every reported metric. PelviNeXt also
outperforms all CNN and transformer baselines trained under our protocol, with
the next-best model, ViT-B/16, trailing by 6.66 percentage points in accuracy
and 0.0710 in AUROC. Notably, the gap between PelviNeXt and prior SOTA is
most pronounced in specificity (87.00\% vs. 82.00\%), suggesting the model is
comparatively better at correctly rejecting normal cases.

\subsection{Ablation Study}
Table~\ref{tab:ablation} reports component-wise ablations on both tasks.
Removing H-CBAM drops accuracy by 1.56 percentage points on PCOS and 2.66
percentage points on fracture, with the larger effect on fracture suggesting
channel-spatial attention is particularly useful for radiographic features.
Removing MSFM causes the largest AUROC drop on both tasks (PCOS: 0.9051
$\to$ 0.8735; fracture: 0.8920 $\to$ 0.8576), identifying multi-scale fusion
as the most critical component for global discriminative performance. Replacing
TH-MHSA with vanilla MHSA yields a smaller but consistent drop (PCOS: 0.9051
$\to$ 0.8851; fracture: 0.8920 $\to$ 0.8860), confirming that head-mixing
provides additional gains beyond standard self-attention. Full PelviNeXt
achieves the best performance on all metrics across both tasks.

\begin{table}[t]
\footnotesize
\setlength{\tabcolsep}{2.5pt}
\renewcommand{\arraystretch}{0.92}
\centering
\caption{Ablation study. Mean $\pm$ 95\% CI over 5-fold CV.}
\label{tab:ablation}
\begin{tabular*}{\columnwidth}{@{\extracolsep{\fill}}lccc@{}}
\hline
& \textbf{Acc.(\%)} & \textbf{F1} & \textbf{AUROC} \\
\multicolumn{4}{l}{\textit{PCOS}} \\
\hline
w/o H-CBAM & 90.44\%$\pm$1.48 & 0.8578$\pm$0.0275 & 0.8758$\pm$0.0196 \\
w/o MSFM & 88.89\%$\pm$1.19 & 0.8509$\pm$0.0190 & 0.8735$\pm$0.0167 \\
Vanilla MHSA & 91.11\%$\pm$2.75 & 0.8692$\pm$0.0317 & 0.8851$\pm$0.0238 \\
\textbf{PelviNeXt} & \textbf{92.00\%$\pm$1.60} & \textbf{0.8890$\pm$0.0085} &
\textbf{0.9051$\pm$0.0156} \\
\hline
\multicolumn{4}{l}{\textit{Fracture}} \\
\hline
w/o H-CBAM & 84.67\%$\pm$3.33 & 0.8579$\pm$0.0270 & 0.8764$\pm$0.0225 \\
w/o MSFM & 83.33\%$\pm$2.92 & 0.8343$\pm$0.0193 & 0.8576$\pm$0.0368 \\
Vanilla MHSA & 86.00\%$\pm$1.31 & 0.8577$\pm$0.0420 & 0.8860$\pm$0.0243 \\
\textbf{PelviNeXt} & \textbf{87.33\%$\pm$2.45} & \textbf{0.8774$\pm$0.0191} &
\textbf{0.8920$\pm$0.0288} \\
\hline
\end{tabular*}
\end{table}

\subsection{Qualitative Analysis}

\begin{figure}[t]
\centering
\includegraphics[width=\columnwidth]{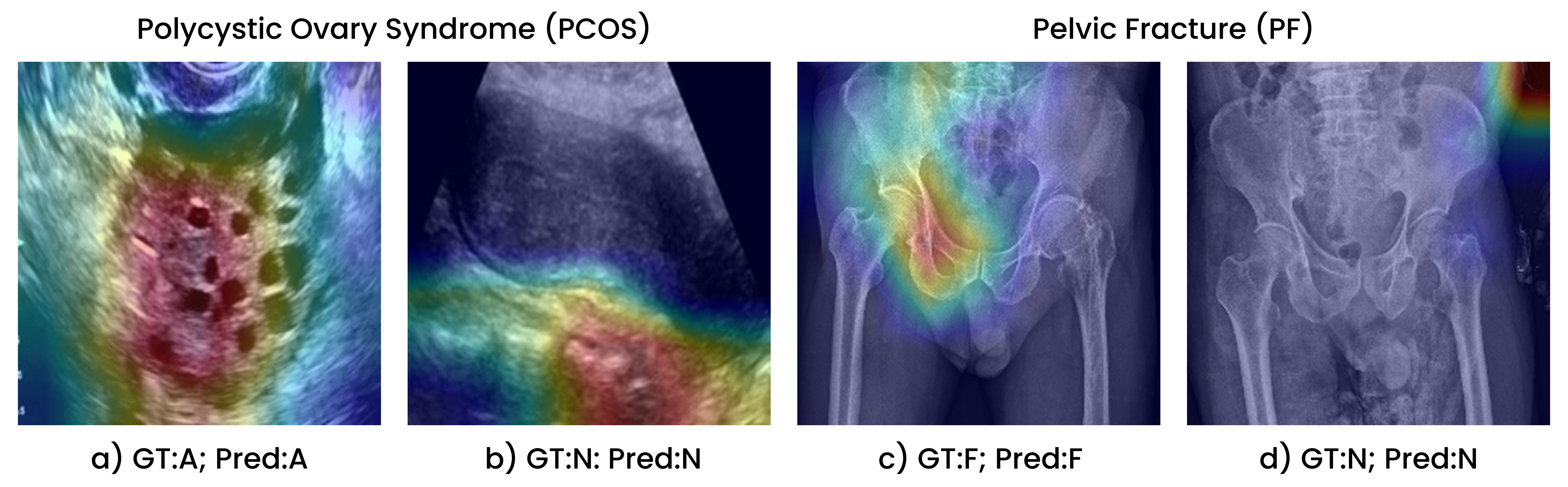}
\caption{Grad-CAM visualizations from PelviNeXt. (a)
Correctly predicted Abnormal PCOS ultrasound. (b) Correctly predicted Normal
PCOS ultrasound. (c) Correctly predicted Fracture pelvic X-ray. (d) Correctly
predicted Normal pelvic X-ray.}
\label{fig:gradcam}
\end{figure}

Figure~\ref{fig:gradcam} presents representative Grad-CAM visualizations \cite{selvaraju2017grad} from both classification tasks. For PCOS ultrasound, the abnormal case exhibits localized activations around the ovarian follicles, whereas the normal case shows more diffuse attention across the ovary. For pelvic fracture radiographs, the fracture case demonstrates concentrated activations within the bone regions, while the normal case produces only weak responses outside the pelvic bones. These qualitative results indicate that PelviNeXt attends to anatomically relevant regions across both imaging modalities.

\subsection{Limitations}
A limitation of this work is the small size of both datasets: the deduplicated
PCOS dataset contains 225 images and PXR150 contains 150 images, which limits
statistical power despite 5-fold CV with 95\% CIs. The PCOS dataset size is a
direct consequence of our integrity audit, and both remain the only publicly
available expert-annotated benchmarks for their respective tasks. Cross-dataset
and cross-site evaluation is important future work once larger public benchmarks
become available.

\section{Conclusion}
We presented PelviNeXt, a modality-agnostic hybrid architecture for pelvic
imaging combining dense feature extraction, H-CBAM, MSFM, and TH-MHSA, applied
without modification to both ultrasound and X-ray. Through a perceptual-hashing
integrity audit of PCOSGen, we identified extensive near-duplicate
contamination, released a deduplicated dataset, and established the first
reliable evaluation baseline. On PXR150, PelviNeXt exceeds prior
state-of-the-art across accuracy, recall, specificity, and AUROC. Ablation
results confirm the contribution of each component. These results take a step
toward reliable, modality-agnostic computer-assisted tools for under-researched
areas of women's pelvic health.

\section*{Data Availability}
The deduplicated PCOSGen benchmark is publicly available at: 

\url{https://www.kaggle.com/datasets/siamtbhuiyan/pcosgen-deduplicated}.

The original PCOSGen and PXR150 datasets are available from their respective
sources~\cite{handa_2025_14592001,handa_2025_14591782,cheng2021scalable}.

\section*{Disclosure of Interests}
The authors have no competing interests to declare.

\bibliographystyle{splncs04}
\bibliography{Paper-016}

@inproceedings{he2016deep,
  title={Deep residual learning for image recognition},
  author={He, Kaiming and Zhang, Xiangyu and Ren, Shaoqing and Sun, Jian},
  booktitle={Proceedings of the IEEE conference on computer vision and pattern recognition},
  pages={770--778},
  year={2016}
}

@article{dosovitskiy2020image,
  title={An image is worth 16x16 words: Transformers for image recognition at scale},
  author={Dosovitskiy, Alexey and Beyer, Lucas and Kolesnikov, Alexander and Weissenborn, Dirk and Zhai, Xiaohua and Unterthiner, Thomas and Dehghani, Mostafa and Minderer, Matthias and Heigold, Georg and Gelly, Sylvain and others},
  journal={arXiv preprint arXiv:2010.11929},
  year={2020}
}

@inproceedings{selvaraju2017grad,
  title={Grad-cam: Visual explanations from deep networks via gradient-based localization},
  author={Selvaraju, Ramprasaath R and Cogswell, Michael and Das, Abhishek and Vedantam, Ramakrishna and Parikh, Devi and Batra, Dhruv},
  booktitle={Proceedings of the IEEE international conference on computer vision},
  pages={618--626},
  year={2017}
}

@article{handa2024auto,
  title={Auto-pcos classification challenge},
  author={Handa, Palak and Saini, Anushka and Dutta, Siddhant and Pathak, Harsh and Choudhary, Nishi and Goel, Nidhi and Dhanao, Jasdeep Kaur and Handa, Palak},
  year={2024},
  publisher={Authorea}
}

@inproceedings{bhuiyan2025patch,
  title={Patch-Based Deep Ensemble Learning for Enhancing Pelvic Fracture Detection},
  author={Bhuiyan, Siam Tahsin and Rahman, Rashedur and Alam, Saadia Binte and Khatun, Halima and Islam, Riyadul and Wasi, Sefatul and Iftee, Md Akil Raihan and Islam, Ashraful},
  booktitle={2025 28th International Conference on Computer and Information Technology (ICCIT)},
  pages={3022--3027},
  year={2025},
  organization={IEEE}
}

@inproceedings{bhuiyan2025preprocessing,
  title={Preprocessing Matters: Benchmarking Image Enhancement Techniques for Pelvic Fracture Detection},
  author={Bhuiyan, Siam Tahsin and Khatun, Rubaya and Mazumder, Samiul Karim and Israq, Fatin and Wasi, Sefatul and Rahman, Rashedur and Islam, Ashraful and Alam, Saadia Binte},
  booktitle={2025 IEEE International Women in Engineering (WIE) Conference on Electrical and Computer Engineering (WIECON-ECE)},
  pages={379--384},
  year={2025},
  organization={IEEE}
}

@article{kassem2023explainable,
  title={Explainable Transfer Learning-Based Deep Learning Model for Pelvis Fracture Detection},
  author={Kassem, Mohamed A and Naguib, Soaad M and Hamza, Hanaa M and Fouda, Mostafa M and Saleh, Mohamed K and Hosny, Khalid M},
  journal={International Journal of Intelligent Systems},
  volume={2023},
  number={1},
  pages={3281998},
  year={2023},
  publisher={Wiley Online Library}
}

@article{shazeer2020talking,
  title={Talking-heads attention},
  author={Shazeer, Noam and Lan, Zhenzhong and Cheng, Youlong and Ding, Nan and Hou, Le},
  journal={arXiv preprint arXiv:2003.02436},
  year={2020}
}

@inproceedings{woo2018cbam,
  title={Cbam: Convolutional block attention module},
  author={Woo, Sanghyun and Park, Jongchan and Lee, Joon-Young and Kweon, In So},
  booktitle={Proceedings of the European conference on computer vision (ECCV)},
  pages={3--19},
  year={2018}
}

@inproceedings{huang2017densely,
  title={Densely connected convolutional networks},
  author={Huang, Gao and Liu, Zhuang and Van Der Maaten, Laurens and Weinberger, Kilian Q},
  booktitle={Proceedings of the IEEE conference on computer vision and pattern recognition},
  pages={4700--4708},
  year={2017}
}

@article{cheng2021scalable,
  title={A scalable physician-level deep learning algorithm detects universal trauma on pelvic radiographs},
  author={Cheng, Chi-Tung and Wang, Yirui and Chen, Huan-Wu and Hsiao, Po-Meng and Yeh, Chun-Nan and Hsieh, Chi-Hsun and Miao, Shun and Xiao, Jing and Liao, Chien-Hung and Lu, Le},
  journal={Nature communications},
  volume={12},
  number={1},
  pages={1066},
  year={2021},
  publisher={Nature Publishing Group UK London}
}

@article{moral2024cystnet,
  title={CystNet: An AI driven model for PCOS detection using multilevel thresholding of ultrasound images},
  author={Moral, Poonam and Mustafi, Debjani and Mustafi, Abhijit and Sahana, Sudip Kumar},
  journal={Scientific reports},
  volume={14},
  number={1},
  pages={25012},
  year={2024},
  publisher={Nature Publishing Group UK London}
}

@dataset{handa_2025_14592001,
  author       = {Handa, Palak and
                  Saini, Anushka and
                  Dutta, Siddhant and
                  Pathak, Harsh and
                  Choudhary, Nishi and
                  Goel, Nidhi and
                  Dhanao, Jasdeep Kaur},
  title        = {PCOSGen-train dataset},
  year         = 2025,
  publisher    = {Zenodo},
  doi          = {10.5281/zenodo.14592001},
  url          = {https://doi.org/10.5281/zenodo.14592001},
}

@dataset{handa_2025_14591782,
  author       = {Handa, Palak and
                  Saini, Anushka and
                  Dutta, Siddhant and
                  Pathak, Harsh and
                  Choudhary, Nishi and
                  Goel, Nidhi and
                  Dhanao, Jasdeep Kaur},
  title        = {PCOSGen-test dataset},
  year         = 2025,
  publisher    = {Zenodo},
  doi          = {10.5281/zenodo.14591782},
  url          = {https://doi.org/10.5281/zenodo.14591782},
}

@article{sang2026benchmark,
  title={Benchmark of Segmentation Techniques for Pelvic Fracture in CT and X-ray: Summary of the PENGWIN 2024 Challenge},
  author={Sang, Yudi and Liu, Yanzhen and Yibulayimu, Sutuke and Wang, Yunning and Killeen, Benjamin D and Liu, Mingxu and Ku, Ping-Cheng and Johannsen, Ole and Gotkowski, Karol and Zenk, Maximilian and others},
  journal={IEEE Transactions on Medical Imaging},
  year={2026},
  publisher={IEEE}
}

@article{lv2022deep,
  title={Deep learning algorithm for automated detection of polycystic ovary syndrome using scleral images},
  author={Lv, Wenqi and Song, Ying and Fu, Rongxin and Lin, Xue and Su, Ya and Jin, Xiangyu and Yang, Han and Shan, Xiaohui and Du, Wenli and Huang, Qin and others},
  journal={Frontiers in Endocrinology},
  volume={12},
  pages={789878},
  year={2022},
  publisher={Frontiers Media SA}
}

@article{bode2026prescribing,
  title={Prescribing policy change to transform women's health research},
  author={Bode, Anna and Fitzgerald, Emily},
  journal={The Lancet Obstetrics, Gynaecology, \& Women’s Health},
  year={2026},
  publisher={Elsevier}
}

@article{moley2024closing,
  title={Closing the gender health gap is a \$1 trillion opportunity},
  author={Moley, Kelle},
  journal={Biopharma Dealmakers},
  year={2024}
}

\end{document}